\documentclass[11pt]{article}
\usepackage[T1]{fontenc}
\usepackage{framed}
\usepackage{acl2016}
\usepackage{times}
\usepackage{url}
\usepackage{latexsym}
\usepackage{graphicx}
\usepackage{amssymb}
\usepackage{subcaption}

\aclfinalcopy 

\title{End-to-end LSTM-based dialog control optimized with \\ supervised and reinforcement learning}

\author{Jason D. Williams and Geoffrey Zweig \\
  Microsoft Research \\
  One Microsoft Way, Redmond, WA 98052, USA \\
  {\tt \{jason.williams,gzweig\}@microsoft.com} 
  \\}

\date{}

\begin{document}
\maketitle
\begin{abstract}
This paper presents a model for end-to-end learning of task-oriented dialog systems.  The main component of the model is a recurrent neural network (an LSTM), which maps from raw dialog history directly to a distribution over system actions.  The LSTM automatically infers a representation of dialog history, which relieves the system developer of much of the manual feature engineering of dialog state.  In addition, the developer can provide software that expresses business rules and provides access to programmatic APIs, enabling the LSTM to take actions in the real world on behalf of the user.  The LSTM can be optimized using supervised learning (SL), where a domain expert provides example dialogs which the LSTM should imitate; or using reinforcement learning (RL), where the system improves by interacting directly with end users.  Experiments show that SL and RL are complementary: SL alone can derive a reasonable initial policy from a small number of training dialogs; and starting RL optimization with a policy trained with SL substantially accelerates the learning rate of RL.
\end{abstract}

\section{Introduction}

Consider how a person would teach another person to conduct a dialog in a particular domain. For example, how an experienced call center agent would help a new agent get started.  First, the teacher would provide an orientation to what ``agent controls'' are available, such as how to look up a customer's information, as well as a few business rules such as how to confirm a customer's identity, or a confirmation message which must be read before performing a financial transaction.  Second, the student would listen in to a few ``good'' dialogs from the teacher, with the goal of imitating them.  Third, the student would begin taking real calls, and the teacher would listen in, providing corrections where the student made mistakes.  Finally, the teacher would disengage, but the student would continue to improve on their own, through experience.

In this paper, we provide a framework for building and maintaining automated dialog systems -- or ``bots'' -- in a new domain that mirrors this progression.  First, a developer provides the set of actions -- both text actions and API calls -- which a bot can invoke, and \emph{action masking} code that indicates when an action is possible given the dialog so far.  Second, a domain expert -- who need not be a developer or a machine learning expert -- provides a set of example dialogs, which a recurrent neural network learns to imitate.  Third, the bot conducts a few conversations, and the domain expert makes corrections.  Finally, the bot interacts with users at scale, improving automatically based on a weak signal that indicates whether dialogs are successful.

Concretely, this paper presents a model of task-oriented dialog control which combines a trainable recurrent neural network with domain-specific software that encodes business rules and logic, and provides access to arbitrary APIs for actions in the domain, such as ordering a taxi or reserving a table at a restaurant.  The recurrent neural network maps directly from a sequence of user turns (represented by the raw words and extracted entities) to actions, and infers its own representation of state.  As a result, minimal hand-crafting of state is required, and no design of a dialog act taxonomy is necessary.  The neural network is trained both using \emph{supervised} learning where ``good'' dialogs are provided for the neural network to imitate, and using \emph{reinforcement learning} where the bot tries new sequences of actions, and improves based on a weak signal of whole-dialog success.  The neural network can be re-trained in under one second, which means that corrections can be made on-line during a conversation, in real time. 

\begin{figure*}[t]
\begin{center}
\includegraphics[trim = 0mm 130mm 60mm 30mm, scale=0.85]{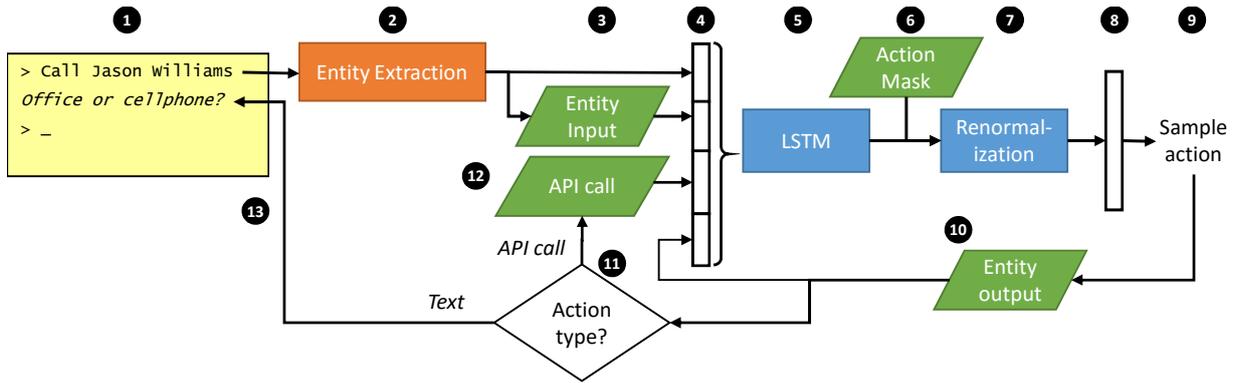}
\caption{\label{fig:diagram}Operational loop.  Green trapezoids refer to programmatic code provided by the software developer.  The blue boxes indicate the recurrent neural network, with trainable parameters.  The orange box performs entity extraction.  The vertical bars in steps 4 and 8 are a feature vector and a distribution over template actions, respectively.  See text for a complete description.}
\end{center}
\end{figure*}


This paper is organized as follows.  First, Section \ref{sec:model} describes the model, and Section \ref{sec:related} compares the model to related work.  Section \ref{sec:app} then presents an example application, which is optimized using supervised learning in Section \ref{sec:sl}, and reinforcement learning in Section \ref{sec:rl}.  Finally, Section \ref{sec:concl} concludes.

\section{Model description}
\label{sec:model}

At a high level, the three components of our model are a recurrent neural network; targeted and well-encapsulated software implementing domain-specific functions; and a language understanding module.  The software enables the developer to express business logic by gating when actions are available; presents a coherent ``surface'' of APIs available to the neural network, such as for placing a phone call; tracks entities which have been mentioned so far in the dialog; and provides features to the neural network which the developer feels may be useful for choosing actions.  The recurrent neural network is responsible for choosing which action to take.  The neural network chooses among action \emph{templates} which abstract over entities, such as the text action ``Do you want to call <name>?'', or the API action \texttt{PlacePhoneCall(<name>)}.  Because a recurrent neural network has internal state, it can accumulate history sufficient for choosing among action templates.  


The components and operational loop are shown in Figure \ref{fig:diagram}.  The cycle begins when the user provides input (step 1).  This input could be text typed in or text recognized from user speech.  This text is passed to an entity extraction module (step 2), which identifies mentions of entities in user text -- for example, identifying ``Jason Williams'' as a <name> entity.  The ``entity input'' (step 3) is code provided by the developer which resolves entity mentions into grounded entities -- in this example, it maps from the text ``Jason Williams'' to a specific row in a database (or a collection of rows in case there are multiple people with this name).  The developer-provided code is stateful, which allows it to retain entities processed in step 3 for use later on in the dialog.  

In step 4, a feature vector is formed, which takes input from 4 sources.  First, the entity extraction module (step 2) indicates which entity types were recognized.  For example, the vector $[1,0]$ could indicate that a name has been recognized, but a type of phone (office vs. mobile) has not.  Second, the entity input module can return arbitrary features specified by the developer.  In this example, this code returns features indicating that ``Jason Williams'' has matched one person, and that ``Jason Williams'' has two types of phones available.  The other two sources are described further below.

Step 5 is a recurrent neural network with a softmax output layer.  In our work, we chose a long short-term memory (LSTM) neural network \cite{lstm} because it has the ability to remember past observations arbitrarily long, and has been shown to yield superior performance in many domains.  The LSTM takes the feature vector from step 4 as input, updates its internal state, and then outputs a distribution over all \emph{template actions} -- i.e., actions with entity values replaced with entity names, as in ``Do you want to call <name>?''.  In step 6, code from the developer outputs an \emph{action mask}, indicating actions which are not permitted at the current timestep.  For example, if a target phone number has not yet been identified, the API action to place a phone call may be masked.\footnote{The action mask is also provided as an input to the LSTM, so it is aware of which actions are currently available; this is not shown in the diagram, for space.}  In step 7, the mask is applied by clamping masked actions to a zero probability, and (linearly) re-normalizing the resulting vector into a probability distribution (step 8).

In step 9, an action is chosen from this probability distribution.  How the action is chosen depends on whether reinforcement learning (RL) is currently active.  When RL is active, exploration is required, so in this case an action is \emph{sampled} from the distribution.  When RL is not active, the best action should be chosen, and so the action with the \emph{highest probability} is always selected.

The identity of the template action selected is then used in 2 ways -- first, it is passed to the LSTM in the next timestep; and second it is passed to the ``entity output'' developer code which substitutes in any template entities.  In step 11, control branches depending on the type of the action: if it is an API text, the corresponding API call in the developer code is invoked (step 12), and any features it returns are passed to the LSTM features in the next timestep.  If the action is text, it is rendered to the user (step 13), and cycle then repeats.


\section{Related work}
\label{sec:related}

In comparing to past work, it is helpful to consider the two main problems that dialog systems solve: \emph{state tracking}, which refers to how information from the past is represented (whether human-interpretable or not), and \emph{action selection}, which refers to how the mapping from state to action is constructed.  We consider each of these in turn.

\subsection{State tracking}

In a task-oriented dialog systems, state tracking typically consists of tracking the \emph{user's goal} such as the cuisine type and price range to use as search criteria for a restaurant, and the \emph{dialog history} such as whether a slot has already been asked for or confirmed, whether a restaurant has been offered already, or whether a user has a favorite cuisine listed in their profile \cite{williams2007csl}.  Most past work to building task-oriented dialog systems has used a \emph{hand-crafted state representation} for both of these quantities -- i.e., the set of possible values for the user's goal and the dialog history are manually designed.  For example, in the Dialog State Tracking Challenge (DSTC), the state consisted of a pre-specified frame of name/value pairs that form the user's goal \cite{dstcoverview}.  Many DSTC entries learned from data how to \emph{update} the state, using methods such as recurrent neural networks \cite{Henderson2014b}, but the \emph{schema} of the state being tracked was hand-crafted.  Manually designed frames are also used for tracking the user's goal and dialog history in methods based on partially observable Markov decision processes (POMDPs) \cite{young2013ieee}, methods which learn from example dialogs \cite{hurtado2005,lee2009example}, supervised learning/reinforcement learning hybrid methods \cite{henderson2005hybrid}, and also in commercial and open source frameworks such as VoiceXML\footnote{\url{www.w3.org/TR/voicexml21}} and AIML.\footnote{\url{www.alicebot.org/aiml.html}}

By contrast, our method \emph{automatically infers a representation of dialog history} in the recurrent neural network which is optimal for predicting actions to take at future timesteps.  This is an important contribution because designing an effective state space can be quite labor intensive: omissions can cause aliasing, and spurious features can slow learning.  Worse, as learning progresses, the set of optimal history features may change.  Thus, the ability to automatically infer a dialog state representation in tandem with dialog policy optimization simplifies developer work.  On the other hand, like past work, the set of possible user goals in our method is hand-crafted -- for many task-oriented systems, this seems desirable in order to support integration with back-end databases, such as a large table of restaurant names, price ranges, etc.  Therefore, our method delegates tracking of user goals to the developer-provided code.\footnote{When entity extraction is reliable -- as it may be in text-based interfaces, which do not have speech recognition errors -- a simple name/value store can track user goals, and this is the approach taken in our example application below.  If entity extraction errors are more prevalent, methods from the dialog state tracking literature for tracking user goals could be applied \cite{dstcoverview}.}

Another line of research has sought to predict the words of the next utterance directly from the history of the dialog, using a recurrent neural network trained on a large corpus of dialogs \cite{lowe2015}.  This work does infer a representation of state; however, our approach differs in several respects: first, in our work, entities are tracked separately -- this allows generalization to entities which have not appeared in the training data; second, our approach includes first-class support for action masking and API calls, which allows the agent to encode business rules and take real-world actions on behalf of the system; finally, in addition to supervised learning, we show how our method can also be trained using reinforcement learning.

\subsection{Action selection}

Broadly speaking, three classes of methods for action selection have been explored in the literature: hand-crafting, supervised learning, and reinforcement learning.  

First, action selection may be hand-crafted, as in VoiceXML, AIML, or a number of long-standing research frameworks \cite{larsson2000information,Seneff2000}.  One benefit of hand-crafted action selection is that business rules can be easily encoded; however, hand-crafting action selection often requires specialized rule engine skills, rules can be difficult to debug, and hand-crafted system don't learn directly from data.

Second, action selection may be learned from example dialogs using supervised learning (SL).  For example, when a user input is received, a corpus of example dialogs can be searched for the most similar user input and dialog state, and the following system action can be output to the user \cite{hurtado2005,lee2009example,4960703,lowe2015,hiraoka2016}.  The benefit of this approach is that the policy can be improved at any time by adding more example dialogs, and in this respect it is rather easy to make corrections in SL-based systems.  However, the system doesn't learn directly from interaction with end users.

Finally, action selection may be learned through reinforcement learning (RL).  In RL, the agent receives a \emph{reward signal} that indicates the quality of an entire dialog, but does not indicate what actions should have been taken.  Action selection via RL was originally framed as a Markov decision process \cite{levin2000stochastic}, and later as a partially observable Markov decision process \cite{young2013ieee}.  If the reward signal naturally occurs, such as whether the user successfully completed a task, then RL has the benefit that it can learn directly from interaction with users, without additional labeling.  Business rules can be incorporated, in a similar manner to our approach \cite{williams2008icslp1}.  However, debugging an RL system is very difficult -- corrections are made via the reward signal, which many designers are unfamiliar with, and which can have non-obvious effects on the resulting policy.  In addition, in early stages of learning, RL performance tends to be quite poor, requiring the use of practice users like crowd-workers or simulated users.   

In contrast to existing work, the neural network in our method can be optimized using \emph{both} supervised learning and reinforcement learning: the neural network is trained using gradient descent, and optimizing with SL or RL simply requires a different gradient computation.  To get started, the designer provides a set of training dialogs, and the recurrent neural network is trained to reconstruct these using supervised learning (Section \ref{sec:sl}).  This avoids poor out-of-the-box performance.  The same neural network can then be optimized using a reward signal, via a policy gradient (Section \ref{sec:rl}).  As with SL-based approaches, if a bug is found, more training dialogs can be added to the training set, so the system remains easy to debug.  In addition, our implementation of RL ensures that the policy always reconstructs the provided training set, so RL optimization will not contradict the training dialogs provided by the designer.  Finally, the action mask provided by the developer code allows business rules to be encoded.

Past work has explored an alternate way of combining supervised learning and reinforcement learning for learning dialog control \cite{henderson2005hybrid}.  In that work, the goal was to learn from a fixed corpus with heterogeneous control policies -- i.e., a corpus of dialogs from many different experts.  The reward function was augmented to penalize policies that deviated from policies found in the corpus.  Our action selection differs in that we view the training corpus as being authoritative -- our goal is to avoid any deviations from the training corpus, and to use RL on-line to improve performance where the example dialogs provide insufficient coverage.

In summary, to our knowledge, this is the first end-to-end method for dialog control which can be trained with both supervised learning and reinforcement learning, and which automatically infers a representation of dialog history while also explicitly tracking entities.  

\section{Example dialog task}
\label{sec:app}

To test our approach, we created a dialog system for initiating phone calls to a contact in an address book, taken from the Microsoft internal employee directory.  In this system, a contact's name may have synonyms (``Michael'' may also be called ``Mike''), and a contact may have more than one phone number, such as ``work'', ``mobile'', etc.  These phone types have synonyms like ``cell'' for ``mobile''.  

We started by defining entities.  The user can say entities <name>, <phonetype>, and <yesno>.  The system can also say these entities, plus three more: <canonicalname> and <canonicalphonetype> allow the user to say a name as in ``call Hillary'' and the system to respond with a canonical name as in ``calling Hillary Clinton''; and <phonetypesavail> which allows the system to say ``Which type of phone: mobile or work?''.  For entity extraction, we trained a model using the Language Understanding Intelligent Service \cite{williams2015luis}.

Next we wrote the programmatic portion of the system.  First, for tracking entities, we used a simple approach where an entity is retained indefinitely after it is recognized, and replaced if a new value is observed.  Then we defined two API actions: one API places a call, and the other commits to a phone type when a contact has only one phone type in the address book.  We then defined features that the back-end can return to the LSTM, including how many people match the most recently recognized name, and how many phone types that person has in the database.  Altogether, the dimension of the LSTM input was $112$ (step 4, Figure \ref{fig:diagram}).  Finally, for the action mask, we allow any action for which the system has all entities -- so ``How can I help you?'' is always available, but the language action ``Calling <name>, <phonetype>'' is only available when the back-end is able to populate those two entities.  Altogether, the code comprised 209 lines of Python.

We then wrote 21 example dialogs, covering scenarios such as when a spoken name has a single vs. multiple address book matches; when there are one vs. more than one phone types available; when the user specifies a phone type and when not; when the user's specified phone type is not available; etc.  One example is given in Figure \ref{fig:example_dialog}, and several more are given in Appendix \ref{app:example_dialogs}.  The example dialogs had on average $7.0$ turns; the longest was $11$ turns and the shortest was $4$ turns.  There were $14$ action templates (step 8, Figure \ref{fig:diagram}).

\begin{figure}[t]
\begin{framed}
\sffamily
\noindent How can I help you?

\noindent \emph{Call Jason}

\noindent Which type of phone: mobile or work?

\noindent \emph{Oh, actually call Mike on his office phone}

\noindent Calling Michael Seltzer, work.






\noindent \texttt{PlaceCall}
\end{framed}
\caption{\label{fig:example_dialog}One of the 21 example dialogs used for supervised learning training.  For space, the entity tags that appear in the user and system sides of the dialogs have been removed -- for example, {\sffamily \emph{Call <name>Jason</name>}} is shown as {\sffamily \emph{Call Jason}}.  See Appendix \ref{app:example_dialogs} for additional examples.}
\end{figure}

In some of the experiments below, we make use of a hand-designed stochastic simulated user.  At the start of a dialog, the simulated user randomly selected a name and phone type, including names and phone types not covered by the dialog system.  When speaking, the simulated user can use the canonical name or a nickname; usually answers questions but can ignore the system; can provide additional information not requested; and can give up.  The simulated user was parameterized by around 10 probabilities, and consisted of 314 lines of Python.

For the LSTM, we selected 32 hidden units, and initialized forget gates to zero, as suggested in \cite{jozefowicz2015empirical}.  The LSTM was implemented using Keras and Theano \cite{chollet2015keras,2016arXiv160502688short}.

\section{Optimizing with supervised learning}
\label{sec:sl}

\subsection{Prediction accuracy}

We first sought to measure whether the LSTM trained with a small number of dialogs would successfully generalize, using a 21-fold leave-one-out cross validation experiment.  In each folds, one dialog was used as the test set, and four different training sets were formed consisting of 1, 2, 5, 10, and 20 dialogs.  Within each fold, a model was trained on each training set then evaluated on the held out test dialog. 

Training was performed using categorical cross entropy as the loss, and with AdaDelta to smooth updates \cite{DBLP:journals/corr/abs-1212-5701}.  Training was run until the training set was reconstructed.

\begin{figure}[t]
\begin{center}
\includegraphics[trim = 20mm 20mm 20mm 20mm, width=\columnwidth]{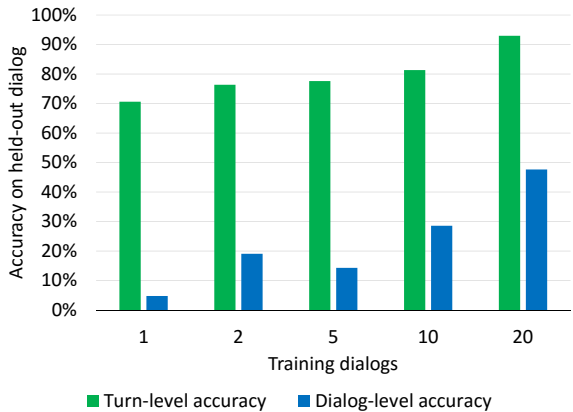}
\caption{\label{fig:testacc}Average accuracy of leave-one-out cross-fold validation.  The $x$ axis shows the number of training dialogs used to train the LSTM.  The $y$ axis shows average accuracy on the one held-out dialog, where green bars show average accuracy measured per turn, and blue bars show average accuracy per dialog.  A dialog is considered accurate if it contains zero prediction errors.}
\end{center}
\end{figure}

Figure \ref{fig:testacc} shows per-turn accuracy and whole-dialog accuracy, averaged across all 21 folds.  After a single dialog, 70\% of dialog turns are correctly predicted.  After 20 dialogs, this rises to over 90\%, with nearly 50\% of dialogs predicted completely correctly.  While this is not sufficient for deploying a final system, this shows that the LSTM is generalizing well enough for preliminary testing after a small number of dialogs.

\subsection{Benefit of recurrency}

We next investigated whether the recurrency in the LSTM was beneficial, or whether a non-stateful deep neural network (DNN) would perform as well.  We substituted the (stateful) LSTM with a non-stateful DNN, with the same number of hidden units as the LSTM, loss function, and gradient accumulator.  We also ran the same experiment with a standard recurrent neural network (RNN).  Training was run until either the training set was reconstructed, or until the loss plateaued for 100 epochs.  Results are shown in Table \ref{fig:dnn}, which shows that the DNN was unable to reconstruct a training set with all 20 dialogs.  Upon investigation, we found that some turns with different actions had identical \emph{local} features, but different histories. Since the DNN is unable to store history, these differences are indistinguishable to the DNN.\footnote{Of course it would be possible to hand-craft additional state features that encode the history, but our goal is to avoid hand-crafting the dialog state as much as possible.}  The RNN also reconstructed the training set; this suggests a line of future work to investigate the relative benefits of different recurrent neural network architectures for this task.

\begin{table}[h]
\begin{center}
\begin{tabular}{|c|ccc|}
\hline \bf Training dialogs & \bf DNN & \bf RNN & \bf LSTM \\ \hline
1  & \checkmark & \checkmark & \checkmark \\
10 & \checkmark & \checkmark & \checkmark \\
21 & $\times$ & \checkmark & \checkmark \\
\hline
\end{tabular}
\end{center}
\caption{\label{fig:dnn} Whether a DNN, RNN and LSTM can reproduce a training set composed of 1, 10, and all 21 training dialogs.  
}
\end{table}

\subsection{Active learning}

\begin{figure}[t]
\begin{center}
\includegraphics[trim = 20mm 20mm 20mm 20mm, width=\columnwidth]{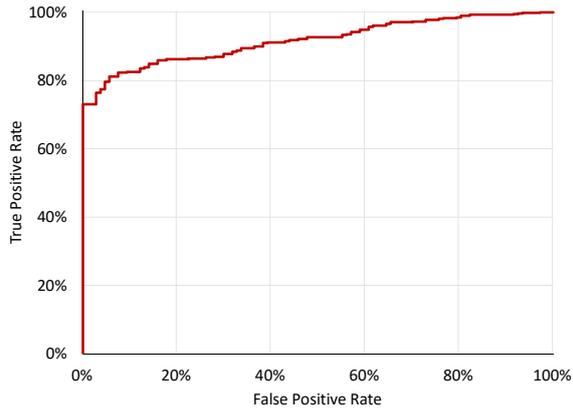}
\caption{\label{fig:roc}ROC plot of the scores of the actions selected by the LSTM.  False positive rate is the number of incorrectly predicted actions above a threshold $r$ divided by the total number of incorrectly predicted actions; true positive rate (TPR) is the number of correctly predicted actions above the threshold $r$ divided by the total number of correctly predicted actions. 
}
\end{center}
\end{figure}

We next examined whether the model would be suitable for active learning \cite{cohn1994improving}.  The goal of active learning is to reduce the number of labels required to reach a given level of performance.  In active learning, the current model is run on (as yet) unlabeled instances, and the unlabeled instances for which the model is most uncertain are labeled next.  The model is then re-built and the cycle repeats.  For active learning to be effective, the scores output by the model must be a good indicator of correctness.  To assess this, we plotted a receiver operating characteristic (ROC) curve, in Figure \ref{fig:roc}.  In this figure, 20 dialogs were randomly assigned to a training set of 11 dialogs and a test set of 10 dialogs.  The LSTM was then estimated on the training set, and then applied to the test set, logging the highest scoring action and that action's correctness.  This whole process was repeated 10 times, resulting in 590 correctly predicted actions and 107 incorrectly predicted actions.   

This figure shows that the model scores are strong predictors of correctness.  
Looking at the lowest scored actions, although incorrectly predicted actions make up just 15\% of turns (107/(590+107)), 80\% of the 20 actions with the lowest scores are incorrect, so labeling low-scoring actions will rapidly correct errors.  

Finally, we note that re-training the LSTM requires less than 1 second on a standard PC (without a GPU), which means the LSTM could be retrained frequently.
Taken together, the model building speed combined with the ability to reliably identify actions which are errors suggests our approach will readily support active learning.

\section{Optimizing with reinforcement learning}
\label{sec:rl}

\begin{figure*}[t]
  \centering
  \begin{subfigure}[b]{0.5\textwidth}
    \centering
    \includegraphics[trim = 20mm 20mm 20mm 20mm, width=\columnwidth]{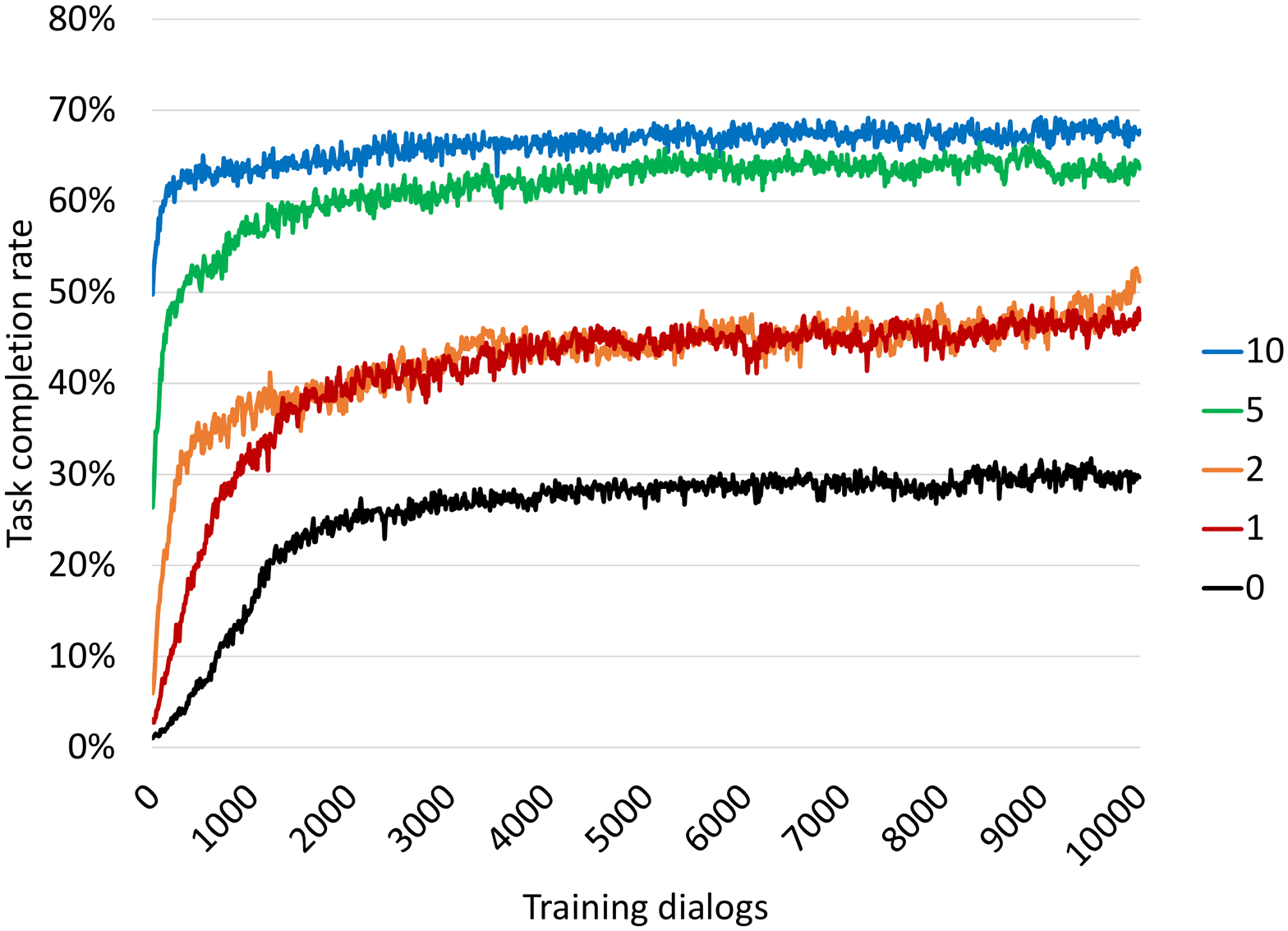}
    \caption{\label{fig:rlperfmean}TCR mean.}
  \end{subfigure}%
  ~
  \begin{subfigure}[b]{0.5\textwidth}
    \centering
    \includegraphics[trim = 20mm 20mm 20mm 20mm, width=\columnwidth]{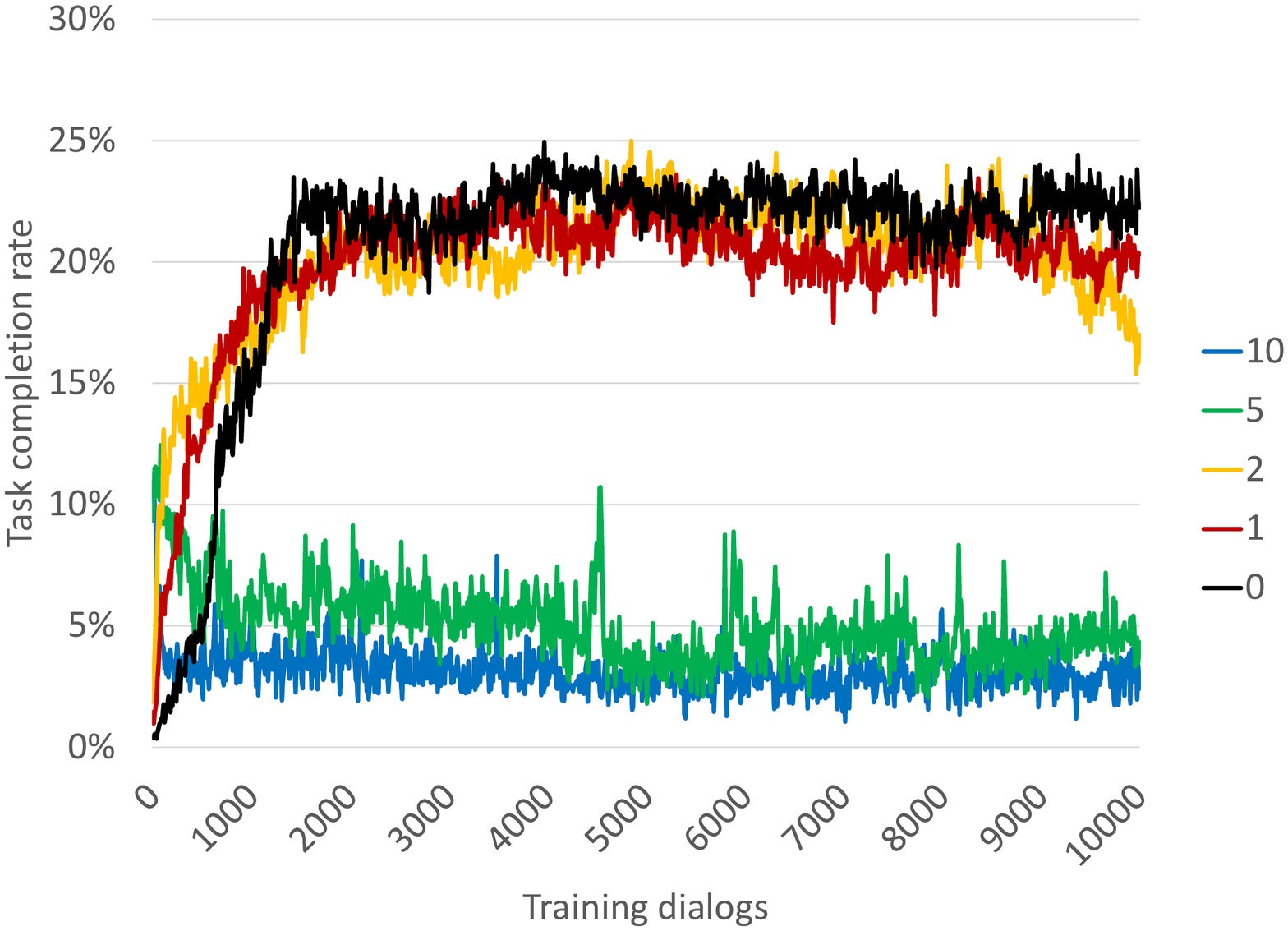}
    \caption{\label{fig:rlperfstddev}TCR standard deviation.}
  \end{subfigure}
  \caption{\label{fig:rlperf}Task completion rate (TCR) mean and standard deviation for a policy initially trained with $N = (0, 1, 2, 5, 10)$ dialogs using supervised learning (SL), and then optimized with $0$ to $10,000$ dialogs using reinforcement learning (RL).  Training and evaluation were done with the same stochastic simulated user.  Each line shows the average of 10 runs, where the dialogs used in the SL training in each run were randomly sampled from the 21 example dialogs.}
\end{figure*}

In the previous sections, supervised learning (SL) was applied to train the LSTM to mimic dialogs provided by the system developer.  Once a system operates at scale, interacting with a large number of users, it is desirable for the system to continue to learn \emph{autonomously} using reinforcement learning (RL).  With RL, each turn receives a measurement of goodness called a \emph{reward}; the agent explores different sequences of actions in different situations, and makes adjustments so as to maximize the expected discounted sum of rewards, which is called the \emph{return}.  We defined the reward as being $1$ for successfully completing the task, and $0$ otherwise.  A discount of $0.95$ was used to incentivize the system to complete dialogs faster rather than slower.

For optimization, we selected a \emph{policy gradient} approach \cite{williams1992simple}.  Conceptually, in policy gradient-based RL, a model outputs a distribution from which actions are sampled at each timestep.  At the end of a dialog, the return for that dialog is computed, and the gradients of the probabilities of the actions taken with respect to the model weights are computed.  The weights are then adjusted by taking a gradient step, weighted by the difference between the return of this dialog and the long-run average return.  Intuitively, ``better'' dialogs receive a positive gradient step, making the actions selected more likely; and ``worse'' dialogs receive a negative gradient step, making the actions selected less likely.  Policy gradient methods have been successfully applied to dialog systems \cite{jurvcivcek2011natural}, robotics \cite{kohl2004policy}, and the board game Go \cite{silver2016mastering}.

The weights $\mathbf{w}$ are updated as
\begin{equation}
\mathbf{w} \leftarrow \mathbf{w} + \alpha ( \sum_t \triangledown_w \log \pi(a_t|h_t;w) ) ( R - b ) \label{eq:policygrad}
\end{equation}
where $\alpha$ is a learning rate; $a_t$ is the action taken at timestep $t$; $h_t$ is the dialog history at time $t$; $R$ is the return of the dialog; $\triangledown_x F$ denotes the Jacobian of $F$ with respect to $x$; $b$ is a baseline described below; and $\pi(a|h;w)$ is the LSTM -- i.e., a stochastic policy which outputs a distribution over $a$ given a dialog history $h$, parameterized by weights $w$.  The baseline $b$ is an estimate of the average return of the current policy, estimated on the last 100 dialogs using weighted importance sampling.\footnote{The choice of baseline does not affect the long-term convergence of the algorithm (i.e., the bias), but does dramatically affect the speed of convergence (i.e., the variance) \cite{williams1992simple}.}

Past work has applied the so-called \emph{natural} gradient estimate \cite{peters2008natural} to dialog systems \cite{jurvcivcek2011natural}.  The natural gradient is a second-order gradient estimate which has often been shown to converge faster than the standard gradient.  However, computing the natural gradient requires inverting a matrix of model weights, which we found to be intractable for the large numbers of weights found in neural networks.  

To the standard policy gradient update, we make three modifications.  First, the effect of the action mask is to clamp some action probabilities to zero, which causes the logarithm term in the policy gradient update to be undefined.  To solve this, we add a small constant to all action probabilities before applying the update.
Second, it is well-known that neural network convergence can be improved using some form of \emph{momentum} -- i.e., accumulation of gradient steps over multiple turns.  In this problem, we found that using AdaDelta sped up convergence substantially \cite{DBLP:journals/corr/abs-1212-5701}.  Finally, in our setting, we want to ensure that the policy continues to reconstruct the example dialogs provided by the developer.  Therefore, after each RL gradient step, we check whether the updated policy reconstructs the training set.  If not, we run \emph{supervised learning} on the training set until the training set is reconstructed.  Note that this approach allows new training dialogs to be added at any time, whether RL optimization is underway or not.  

We evaluate RL optimization in two ways.  First, we randomly initialize an LSTM, and begin RL optimization.  
Second, we initialize the LSTM by first applying supervised learning on a training set, consisting of 1, 2, 5, or 10 dialogs, formed by randomly sampling from the 21 example dialogs.  RL policy updates are made after each dialog.  After 10 RL updates, we freeze the policy, and run 500 dialogs with the user simulation to measure task completion.  We repeat all of this for 10 runs, and report average performance.

Results are shown in Figure \ref{fig:rlperf}.  RL alone ($N=0$) sometimes fails to discover a complete policy -- in the first 10,000 dialogs, some runs of RL with fewer SL pre-training dialogs failed to discover certain action sequences, resulting in lower average task completion -- for the black line, note the low average in Figure \ref{fig:rlperfmean} and the high variance in Figure \ref{fig:rlperfstddev}.  The difficulty of discovering long action sequences with delayed rewards has been observed in other applications of RL to dialog systems \cite{williams2007ts}.  By contrast, the addition of a few dialogs and pre-training with SL both accelerates learning on average, and reduces the variability in performance of the resulting policy.  

\section{Conclusion}
\label{sec:concl}

This paper has taken a first step toward end-to-end learning of task-oriented dialog systems.  Our approach is based on a recurrent neural network which maps from raw dialog history to distributions over actions.  The LSTM automatically infers a representation of dialog state, alleviating much of the work of hand-crafting a representation of dialog state.  Code provided by the developer tracks entities, wraps API calls to external actuators, and can enforce business rules on the policy.  Experimental results have shown that training with supervised learning yields a reasonable policy from a small number of training dialogs, and that this initial policy accelerates optimization with reinforcement learning substantially.  To our knowledge, this is the first demonstration of end-to-end learning of dialog control for task-oriented domains.

\bibliography{master}

\begin{thebibliography}{}

\bibitem[\protect\citename{Chollet}2015]{chollet2015keras}
François Chollet.
\newblock 2015.
\newblock Keras.
\newblock \url{https://github.com/fchollet/keras}.

\bibitem[\protect\citename{Cohn \bgroup et al.\egroup }1994]{cohn1994improving}
David Cohn, Les Atlas, and Richard Ladner.
\newblock 1994.
\newblock Improving generalization with active learning.
\newblock {\em Machine learning}, 15(2):201--221.

\bibitem[\protect\citename{Henderson \bgroup et al.\egroup
  }2005]{henderson2005hybrid}
James Henderson, Oliver Lemon, and Kalliroi Georgila.
\newblock 2005.
\newblock Hybrid reinforcement/supervised learning for dialogue policies from
  {C}ommunicator data.
\newblock In {\em Proc Workshop on Knowledge and Reasoning in Practical
  Dialogue Systems, Intl Joint Conf on Artificial Intelligence (IJCAI),
  Edinburgh}, pages 68--75.

\bibitem[\protect\citename{Henderson \bgroup et al.\egroup
  }2014]{Henderson2014b}
Matthew Henderson, Blaise Thomson, and Steve Young.
\newblock 2014.
\newblock {Word-based Dialog State Tracking with Recurrent Neural Networks}.
\newblock In {\em Proc SIGdial Workshop on Discourse and Dialogue,
  Philadelphia, USA}.

\bibitem[\protect\citename{Hiraoka \bgroup et al.\egroup }2016]{hiraoka2016}
Takuya Hiraoka, Graham Neubig, Koichiro Yoshino, Tomoki Toda, and Satoshi
  Nakamura.
\newblock 2016.
\newblock Active learning for example-based dialog systems.
\newblock In {\em Proc Intl Workshop on Spoken Dialog Systems, Saariselka,
  Finland}.

\bibitem[\protect\citename{Hochreiter and Schmidhuber}1997]{lstm}
Sepp Hochreiter and Jurgen Schmidhuber.
\newblock 1997.
\newblock Long short-term memory.
\newblock {\em Neural Computation}, 9(8):1735–--1780.

\bibitem[\protect\citename{Hori \bgroup et al.\egroup }2009]{4960703}
Chiori Hori, Kiyonori Ohtake, Teruhisa Misu, Hideki Kashioka, and Satoshi
  Nakamura.
\newblock 2009.
\newblock {Statistical dialog management applied to WFST-based dialog systems}.
\newblock In {\em Acoustics, Speech and Signal Processing, 2009. ICASSP 2009.
  IEEE International Conference on}, pages 4793--4796, April.

\bibitem[\protect\citename{Hurtado \bgroup et al.\egroup }2005]{hurtado2005}
Lluis~F. Hurtado, David Griol, Emilio Sanchis, and Encarna Segarra.
\newblock 2005.
\newblock A stochastic approach to dialog management.
\newblock In {\em Proc IEEE Workshop on Automatic Speech Recognition and
  Understanding (ASRU), San Juan, Puerto Rico, USA}.

\bibitem[\protect\citename{Jozefowicz \bgroup et al.\egroup
  }2015]{jozefowicz2015empirical}
Rafal Jozefowicz, Wojciech Zaremba, and Ilya Sutskever.
\newblock 2015.
\newblock An empirical exploration of recurrent network architectures.
\newblock In {\em Proceedings of the 32nd International Conference on Machine
  Learning (ICML-15)}, pages 2342--2350.

\bibitem[\protect\citename{Jur{\v{c}}{\'\i}{\v{c}}ek \bgroup et al.\egroup
  }2011]{jurvcivcek2011natural}
Filip Jur{\v{c}}{\'\i}{\v{c}}ek, Blaise Thomson, and Steve Young.
\newblock 2011.
\newblock Natural actor and belief critic: Reinforcement algorithm for learning
  parameters of dialogue systems modelled as pomdps.
\newblock {\em ACM Transactions on Speech and Language Processing (TSLP)},
  7(3):6.

\bibitem[\protect\citename{Kohl and Stone}2004]{kohl2004policy}
Nate Kohl and Peter Stone.
\newblock 2004.
\newblock Policy gradient reinforcement learning for fast quadrupedal
  locomotion.
\newblock In {\em Robotics and Automation, 2004. Proceedings. ICRA'04. 2004
  IEEE International Conference on}, volume~3, pages 2619--2624. IEEE.

\bibitem[\protect\citename{Larsson and Traum}2000]{larsson2000information}
Staffan Larsson and David Traum.
\newblock 2000.
\newblock Information state and dialogue management in the {TRINDI} dialogue
  move engine toolkit.
\newblock {\em Natural Language Engineering}, 5(3/4):323--340.

\bibitem[\protect\citename{Lee \bgroup et al.\egroup }2009]{lee2009example}
Cheongjae Lee, Sangkeun Jung, Seokhwan Kim, and Gary~Geunbae Lee.
\newblock 2009.
\newblock Example-based dialog modeling for practical multi-domain dialog
  system.
\newblock {\em Speech Communication}, 51(5):466--484.

\bibitem[\protect\citename{Levin \bgroup et al.\egroup
  }2000]{levin2000stochastic}
Esther Levin, Roberto Pieraccini, and Wieland Eckert.
\newblock 2000.
\newblock A stochastic model of human-machine interaction for learning dialogue
  strategies.
\newblock {\em IEEE Trans on Speech and Audio Processing}, 8(1):11--23.

\bibitem[\protect\citename{Lowe \bgroup et al.\egroup }2015]{lowe2015}
Ryan Lowe, Nissan Pow, Iulian Serban, and Joelle Pineau.
\newblock 2015.
\newblock The ubuntu dialogue corpus: A large dataset for research in
  unstructured multi-turn dialogue systems.
\newblock In {\em Proc SIGdial Workshop on Discourse and Dialogue, Prague,
  Czech Republic}.

\bibitem[\protect\citename{Peters and Schaal}2008]{peters2008natural}
Jan Peters and Stefan Schaal.
\newblock 2008.
\newblock Natural actor-critic.
\newblock {\em Neurocomputing}, 71(7):1180--1190.

\bibitem[\protect\citename{Seneff and Polifroni}2000]{Seneff2000}
Stephanie Seneff and Joseph Polifroni.
\newblock 2000.
\newblock Dialogue management in the mercury flight reservation system.
\newblock In {\em Proceedings of the 2000 ANLP/NAACL Workshop on Conversational
  Systems - Volume 3}, ANLP/NAACL-ConvSyst '00, pages 11--16. Association for
  Computational Linguistics.

\bibitem[\protect\citename{Silver \bgroup et al.\egroup
  }2016]{silver2016mastering}
David Silver, Aja Huang, Chris~J. Maddison, Arthur Guez, Laurent Sifre, George
  Van Den~Driessche, Julian Schrittwieser, Ioannis Antonoglou, Veda
  Panneershelvam, Marc Lanctot, et~al.
\newblock 2016.
\newblock {Mastering the game of Go with deep neural networks and tree search}.
\newblock {\em Nature}, 529(7587):484--489.

\bibitem[\protect\citename{{Theano Development
  Team}}2016]{2016arXiv160502688short}
{Theano Development Team}.
\newblock 2016.
\newblock {Theano: A {Python} framework for fast computation of mathematical
  expressions}.
\newblock {\em arXiv e-prints}, abs/1605.02688, May.

\bibitem[\protect\citename{Williams and Young}2007]{williams2007csl}
Jason~D. Williams and Steve Young.
\newblock 2007.
\newblock Partially observable {M}arkov decision processes for spoken dialog
  systems.
\newblock {\em Computer Speech and Language}, 21(2):393--422.

\bibitem[\protect\citename{Williams \bgroup et al.\egroup
  }2015]{williams2015luis}
Jason~D. Williams, Eslam Kamal, Mokhtar Ashour, Hani Amr, Jessica Miller, and
  Geoff Zweig.
\newblock 2015.
\newblock Fast and easy language understanding for dialog systems with
  microsoft language understanding intelligent service (luis).
\newblock In {\em Proc SIGdial Workshop on Discourse and Dialogue, Prague,
  Czech Republic}.

\bibitem[\protect\citename{Williams \bgroup et al.\egroup }2016]{dstcoverview}
Jason~D. Williams, Antoine Raux, and Matthew Henderson.
\newblock 2016.
\newblock The dialog state tracking challenge series: A review.
\newblock {\em Dialogue and Discourse}, 7(3).

\bibitem[\protect\citename{Williams}1992]{williams1992simple}
Ronald~J Williams.
\newblock 1992.
\newblock Simple statistical gradient-following algorithms for connectionist
  reinforcement learning.
\newblock {\em Machine learning}, 8(3-4):229--256.

\bibitem[\protect\citename{Williams}2007]{williams2007ts}
Jason~D. Williams.
\newblock 2007.
\newblock Applying {POMDP}s to dialog systems in the troubleshooting domain.
\newblock In {\em NAACL-HLT Workshop on Bridging the Gap: Academic and
  Industrial Research in Dialog Technologies, Rochester, New York, USA}, pages
  1--8.

\bibitem[\protect\citename{Williams}2008]{williams2008icslp1}
Jason~D. Williams.
\newblock 2008.
\newblock The best of both worlds: {U}nifying conventional dialog systems and
  {POMDP}s.
\newblock In {\em Proc Intl Conf on Spoken Language Processing (ICSLP),
  Brisbane, Australia}.

\bibitem[\protect\citename{Young \bgroup et al.\egroup }2013]{young2013ieee}
Steve Young, Milica Gasic, Blaise Thomson, and Jason~D. Williams.
\newblock 2013.
\newblock {POMDP-based Statistical Spoken Dialogue Systems: a Review}.
\newblock {\em {Proceedings of the IEEE}}, PP(99):1--20.

\bibitem[\protect\citename{Zeiler}2012]{DBLP:journals/corr/abs-1212-5701}
Matthew~D. Zeiler.
\newblock 2012.
\newblock {ADADELTA:} an adaptive learning rate method.
\newblock {\em CoRR}, abs/1212.5701.

\end{thebibliography}
\bibliographystyle{acl2016}

\appendix

\section{Example dialogs}
\label{app:example_dialogs}
A subset of the 21 example dialogs created for this paper are given below.  For space, the entity tags that appear in the user and system sides of the dialogs have been removed -- for example, {\sffamily \emph{Call <name>Jason</name>}} is shown as {\sffamily \emph{Call Jason}}.

\sffamily
\begin{framed}
\noindent How can I help you?

\noindent \emph{Call Jason Williams cellphone}

\noindent Calling Jason Williams, mobile

\noindent \texttt{PlaceCall}

\end{framed}
\begin{framed}
\noindent How can I help you?

\noindent \emph{Call Jason Williams}

\noindent Which type of phone: mobile or work?

\noindent \emph{office}

\noindent Calling Jason Williams, work

\noindent \texttt{PlaceCall}

\end{framed}
\begin{framed}
\noindent How can I help you?

\noindent \emph{Call Frank on his cellphone}

\noindent Sorry, I don't have a cellphone number for Frank Seide.  I only have a work phone.  Do you want to call that number?

\noindent \emph{yes}

\noindent \texttt{SavePhonetypeavail}

\noindent Calling Frank Seide, work

\noindent \texttt{PlaceCall}

\end{framed}
\begin{framed}
\noindent How can I help you?

\noindent \emph{Call Frank on his home}

\noindent Sorry, I don't have a home number for Frank Seide.  I only have a work phone.  Do you want to call that number?

\noindent \emph{no}

\noindent Oh, sorry about that.  Goodbye.

\end{framed}
\begin{framed}
\noindent How can I help you?

\noindent \emph{Call Jason at his home}

\noindent Sorry, I don't have a home number for Jason Williams.  I have mobile or work.  Which would you like?

\noindent \emph{mobile}

\noindent Calling Jason Williams, mobile

\noindent \texttt{PlaceCall}

\end{framed}
\begin{framed}
\noindent How can I help you?

\noindent \emph{Call Michael}

\noindent Theres more than one person named Michael. Can you say their full name?

\noindent \emph{Call Michael Seltzer}

\noindent \texttt{SavePhonetypeavail}

\noindent Calling Michael Seltzer, work

\noindent \texttt{PlaceCall}

\end{framed}
\begin{framed}
\noindent How can I help you?

\noindent \emph{Call Michel}

\noindent Sorry, I don't know of any names called Michel.  Can you try again?

\noindent \emph{Call Michael Seltzer}

\noindent \texttt{SavePhonetypeavail}

\noindent Calling Michael Seltzer, work

\noindent \texttt{PlaceCall}

\end{framed}

\end{document}